\documentclass[conference]{IEEEtran}
\IEEEoverridecommandlockouts

\usepackage{cite}
\usepackage{amsmath,amssymb,amsfonts}
\usepackage{algorithmic}
\usepackage{graphicx}
\usepackage{textcomp}
\usepackage{xcolor}

\usepackage{subfigure}

\usepackage{booktabs}
\usepackage{multirow}
\usepackage{enumerate}
\usepackage{color,xcolor}
\usepackage{caption}

\def\BibTeX{{\rm B\kern-.05em{\sc i\kern-.025em b}\kern-.08em
    T\kern-.1667em\lower.7ex\hbox{E}\kern-.125emX}}
\begin{document}

\title{Spatio-Temporal Multi-Subgraph GCN for 3D Human Motion Prediction
}


\author{\IEEEauthorblockN{1\textsuperscript{st} Jiexin Wang}
\IEEEauthorblockA{\textit{Gaoling School of Artificial Intelligence} \\
\textit{Renmin University of China}\\
Beijing, China \\
jiexinwang@ruc.edu.cn}
\and
\IEEEauthorblockN{2\textsuperscript{nd} Yiju Guo}
\IEEEauthorblockA{\textit{Gaoling School of Artificial Intelligence} \\
\textit{Renmin University of China}\\
Beijing, China \\
yijuguo@ruc.edu.cn}
\and
\IEEEauthorblockN{3\textsuperscript{rd} Bing Su*
\thanks{* Corresponding author.}
}
\IEEEauthorblockA{\textit{Gaoling School of Artificial Intelligence} \\
\textit{Renmin University of China}\\
Beijing, China \\
subingats@gmail.com}
}

\maketitle

\begin{abstract}
Human motion prediction (HMP) involves forecasting future human motion based on historical data. Graph Convolutional Networks (GCNs) have garnered widespread attention in this field for their proficiency in capturing relationships among joints in human motion. However, existing GCN-based methods tend to focus on either temporal-domain or spatial-domain features, or they combine spatio-temporal features without fully leveraging the complementarity and cross-dependency of these two features. 
In this paper, we propose the Spatial-Temporal Multi-Subgraph Graph Convolutional Network (STMS-GCN) to capture complex spatio-temporal dependencies in human motion. Specifically, we decouple the modeling of temporal and spatial dependencies, enabling cross-domain knowledge transfer at multiple scales through a spatio-temporal information consistency constraint mechanism. Besides, we utilize multiple subgraphs to extract richer motion information and enhance the learning associations of diverse subgraphs through a homogeneous information constraint mechanism. 
Extensive experiments on the standard HMP benchmarks demonstrate the superiority of our method.
\end{abstract}


\section{Introduction}
\label{sec:intro}

3D Skeleton-based human motion prediction (HMP) involves forecasting future motion sequences based on given historical motion sequences. This research is crucial for comprehending human motion behavior and has widespread applications in various domains~\cite{unhelkar2018human,gopalakrishnan2019neural,chen20203d}. 
Advancements in deep learning have significantly improved HMP by enabling the learning of complex spatio-temporal representations. Techniques such as Convolutional Neural Networks (CNN)~\cite{li2018convolutional,liu2020trajectorycnn,cui2021efficient}, Recurrent Neural Networks (RNN)~\cite{martinez2017human,corona2020context}, Long Short-Term Memory (LSTM) ~\cite{tang2018long,lee2021video}, and Transformer Networks~\cite{cai2020learning,martinez2021pose,xu2023auxiliary} have leaded to more accurate and reliable predictions.
However, these methods exhibit certain limitations. For instance, CNN-based approaches heavily rely on filter sizes, while RNN and LSTM-based methods suffer from error accumulation issues. 
Currently, Graph Convolutional Networks (GCNs) have been a mainstream framework for motion prediction due to their exceptional ability to model intrinsic kinematic dependencies and learn spatial relationships among joints in HMP~\cite{li2022skeleton,gao2023decompose,zhang2023dynamic}. 

\begin{figure}[!t]
  \centering
\subfigure[Individual]{\includegraphics[width=0.239\linewidth]{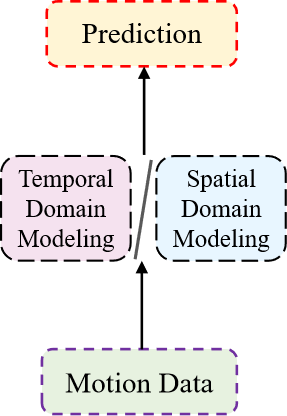}\label{fig:1_1}}
\subfigure[Individual]{\includegraphics[width=0.188\linewidth]{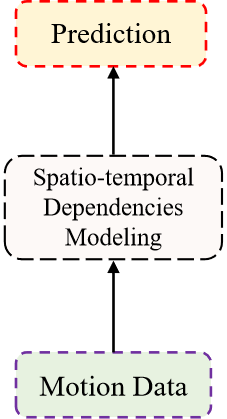}\label{fig:1_2}}
\subfigure[Separated]{\includegraphics[width=0.233\linewidth]{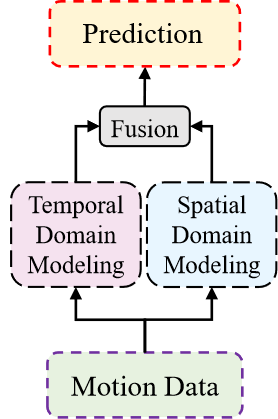}\label{fig:1_3}}
\subfigure[Ours]{\includegraphics[width=0.295\linewidth]{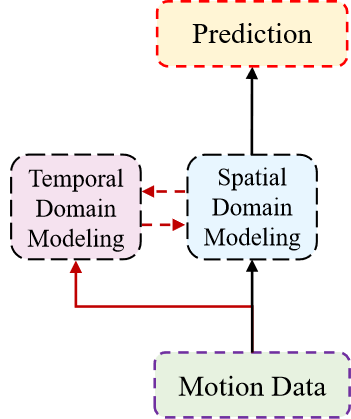}\label{fig:1_4}}
\vspace{-0.10in}
{ \caption{Comparison of prediction methods.
(a): Individual modeling of spatial or temporal dependencies. 
(b): Mixing spatio-temporal dependencies modeling.
(c): Decoupling motion data modeling into temporal and spatial domains, fusing features for final prediction.
(d): Ours leverages temporal-domain learning to assist the learning of the spatial domain, distilling the learned cross-domain knowledge into interactions across multiple scales (red dashed lines).}
\vspace{-0.20in}
\label{fig:method_compare}}
\end{figure}

Most GCN-based methods focus solely on modeling either temporal or spatial features~\cite{liu2019skeleton,li2020dynamic,dang2021msr,li2022skeleton}, as depicted in Fig.\ref{fig:1_1}. Another prevalent approach involves integrating spatio-temporal relationships by blending various convolutional kernels~\cite{cui2021towards,ma2022progressively,feichtenhofer2022masked,tan2023temporal,gao2023decompose}, as shown in Fig.\ref{fig:1_2}. However, the former overlooks the complementarity between temporal and spatial features, while the latter struggles to extract distinct temporal and spatial information, potentially reducing predictive performance.
Besides, some models separately model the temporal and spatial dependencies, then fuse the spatio-temporal features for motion prediction motion~\cite{zhong2022spatio,wang2023spatio}, as seen in Fig.\ref{fig:1_3}. However, they mainly focus on designing extra sophisticated structures to work with the traditional GCN or ignore the hidden cross-dependency in spatio-temporal relationships. Consequently, the complementary nature of temporal and spatial information has not been fully explored.

In this paper, we propose a novel spatio-temporal graph convolutional network, depicted in Fig.\ref{fig:1_4}. Concretely, we employ orthogonal spatio-temporal branches to separately model temporal and spatial domain information, fully leveraging the uniqueness of spatio-temporal information. This design allows each branch to focus solely on learning from one dimension, harnessing the model's understanding of single-modal information. Moreover, taking into account that direct concatenation of the decoupled spatial and temporal features may not fully exploit the complementarity and cross-dependency of the spatio-temporal relationships, we propose a consistent constraint between the intermediate features of the two branches. This constraint facilitates the interweaving of spatio-temporal information and the transfer of cross-domain knowledge, thereby enhancing the model's capacity to effectively capture and utilize the intricate spatio-temporal relationships within human motion.

Additionally, the abundance of motion information is critical for the learning of predictive models~\cite{li2020dynamic,dang2021msr,zhong2022spatio}. Inspired by the Mixture of Experts (MoE)~\cite{eigen2013learning,shazeer2017outrageously}, we introduce multi-subgraph learning to enhance the model's comprehension of the intricate spatio-temporal dependencies. Technically, we incorporate multiple trainable graph convolution operators, each acting as a subgraph, to process the shared motion feature. Different subgraphs offer distinct insights into the shared motion patterns, effectively capturing inherent motion patterns. 
To ensure controlled learning of different subgraphs and prevent excessive divergence in learning between them, we propose a regularizing term designed for the learning of homogeneous information~\cite{liu2023diversifying,lo2024closer}. This above design not only accurately captures complex motion relationships but also enhances the model's generalization performance across diverse types of motion. Through extensive experiments and comparisons with current GCN-based approaches, we demonstrate the effectiveness and superiority of our method.
The contributions of this paper are summarized as follows:
\begin{enumerate}[1)]
\item We propose the spatio-temporal multi-subgraph graph convolutional network (STMS-GCN) for human motion prediction, which considers the independence and complementary of spatio-temporal information, utilizing diverse learnable subgraphs to capture richer motion patterns.
\item We propose a cross-domain information contrast mechanism to enhance the interaction between spatial and temporal information across multiple scales. Moreover, we propose a homogeneous information constraint mechanism to meticulously regulate subgraph learning.
\item We conduct extensive experiments on the standard HMP benchmarks. The experimental results demonstrate the effectiveness and superiority of our method in accurately predicting human motion across diverse scenarios.
\end{enumerate}
\section{Methodology}
\label{Methodology}

\subsection{Problem Formulation}

Let $\mathbf{X}=[X_1,\cdots,X_{T_p}] \in \mathbb{R}^{T_p \times J \times D}$ represent the given historical poses, and $\mathbf{Y} = [X_{T_p+1},\cdots,X_{T_p+T_f}] \in \mathbb{R}^{T_f \times J \times D}$ denote the predicted motion sequence for the next $T_f$ time steps. Each pose $X_{t} \in \mathbb{R}^{J \times D}$ describes a human pose with $J$ joints in $D$ dimensions at time $t$. Typically, $D$ is equal to 2 or 3, indicating the 2D or 3D case.
Following~\cite{mao2019learning,mao2020history}, we pad the last observed pose $X_{T_p}$ by repeating it $T_f$ times and append the resulting poses to $\mathbf{X}$. Then we have the padded input $\mathbf{X}= [X_1,\cdots,X_{T_p},X_{T_p},\cdots,X_{T_p}] \in \mathbb{R}^{(T_p+T_f) \times J \times D}$. Correspondingly, we construct $\mathbf{Y} = [X_1,\cdots,X_{T_p+T_f}] \in \mathbb{R}^{(T_p+T_f) \times J \times D}$.
Formally, the problem is represented as:
\begin{equation}
\widetilde{\mathbf{Y}} = \mathcal{F}_\mathrm{{pred}}(\mathbf{X}), 
\label{EQ:objective}
\end{equation}
where $\mathcal{F}_\mathrm{{pred}}$ denotes a predictor, which bridges the past motion $\mathbf{X}$ to future motion $\mathbf{Y}$. 
The aim of HMP is that the prediction $\widetilde{\mathbf{Y}}$ is as accurate as possible compared to $\mathbf{Y}$. 

\begin{figure}[!t]
    \centering
    \includegraphics[width=0.95\linewidth]{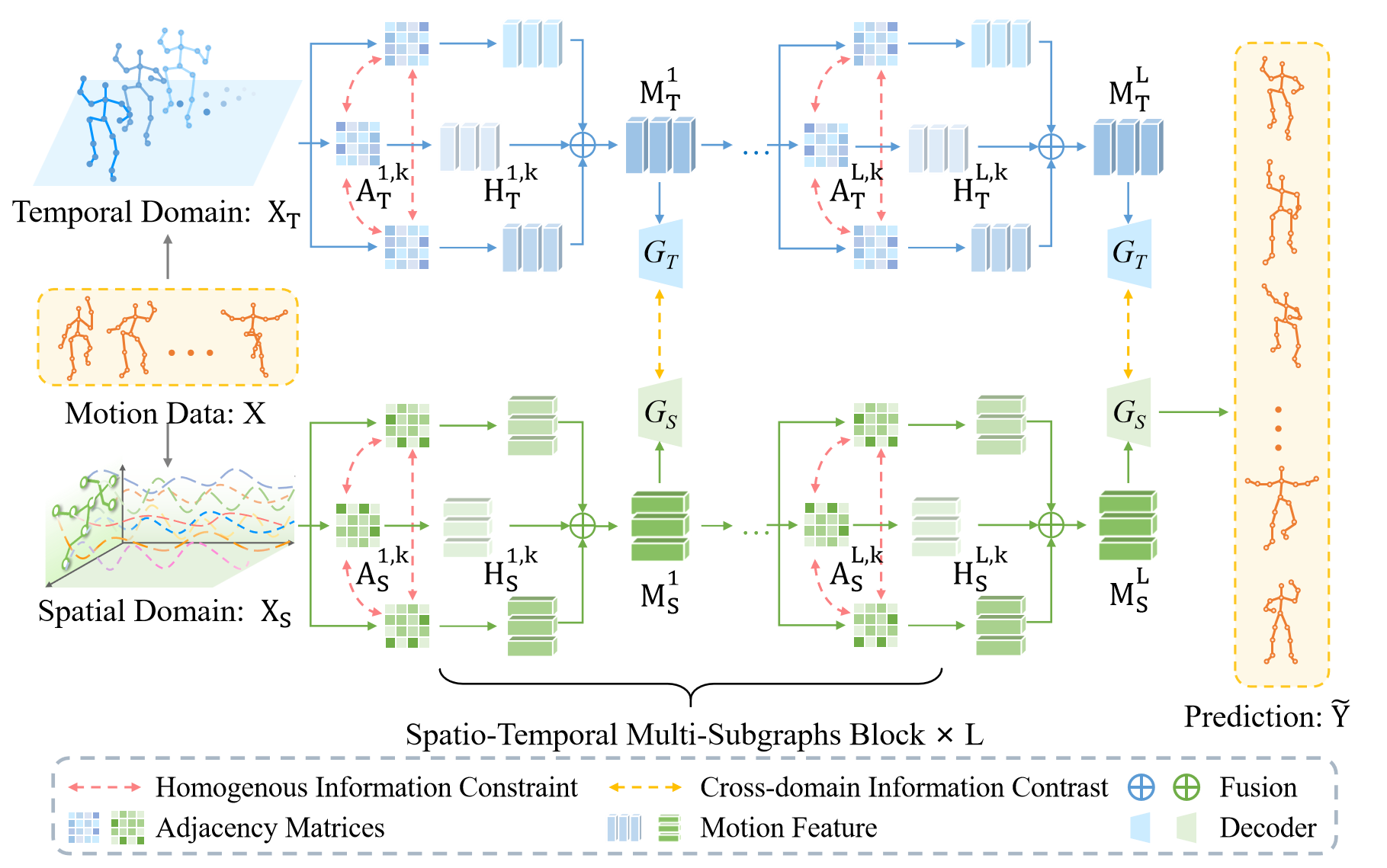}
    \vspace{-0.1in}
    \caption{Illustration of STMS-GCN. 
    }
    \label{fig:method}
    \vspace{-0.20in}
\end{figure}

\subsection{Spatio-Temporal Multi-Subgraph Block}
Current GCN-based methods suffer limitations by predominantly focusing on either temporal or spatial features, disregarding the potential synergy between them~\cite{luo2018fast,li2021skeleton}. 
Despite some methods attempt to capture spatio-temporal relationships through a combination of convolutional kernels, they neglect the inherent complementarity between these features~\cite{hernandez2019human,ma2022progressively,zhang2023dynamic}.
To tackle this, we advocate the use of two orthogonal spatio-temporal branches that complement and mutually reinforce each other.
We propose the Spatio-Temporal Multi-Subgraph Block (STMSB), as shown in~\figurename{~\ref{fig:method}}, which consists of two critical modules: 1) the spatio-temporal double branch and 2) multi-subgraph learning. The STMSB forms a network with $L$ layers. Here we consider the $l$-th layer as an example, and we could expand the design to any layers. In the following sections, we introduce these modules in detail.

\subsection{Spatio-temporal Double Branch}
Given $\mathbf{X}$ as the input, we separately encode $\mathbf{X}$ from the temporal and spatial domain to model the temporal smoothness and spatial dependencies among human body joints.

\noindent\textbf{Temporal Domain Modeling.}
We reshape $\mathbf{X}$ into $\mathbf{X}_T = \{ X_{T,i}\}_{i=1}^{T_p+T_f}\in \mathbb{R}^{(T_p+T_f) \times J \cdot D}$ as the input to the temporal branch. Then, a frame embedding is employed to project $\mathbf{X}_T$ into a $C$-dimensional dense feature space through two fully connected layers, which can be formalized as: 
\begin{equation}
    \hat{X}_{T,i} = W_{2} \cdot ( \sigma ( W_{1} \cdot X_{T,i} + b_{1}) ) + b_{2},
    \label{Eq:embedding_t}
\end{equation}
where $ W_{1}\in \mathbb{R}^{C \times J \cdot D}$ and $ W_{2}\in \mathbb{R}^{C \times C}$ are transformation matrices, $b_{1}\in \mathbb{R}^{C}$ and $b_{2}\in \mathbb{R}^{C} $ indicate the bias vectors, and $\sigma$ is the ReLU function. All projected features are jointly denoted as $\hat{\mathbf{X}}_T =\{\hat {X}_{T,i} \}_{i=1}^{T_p+T_f}\in \mathbb{R}^{(T_p+T_f) \times C}$. 
To capture temporal dependencies between the independent frames in $\hat{\mathbf{X}}_T$, we employ GCN, treating each frame as a node in the graph to capture relationships across different frames. 
\begin{equation}
\left \{
\begin{array}{lll}
l=1 : \mathbf{M}_T^l = \mathbb{f}_T^l(\hat{\mathbf{X}}_T) \\
l>1 : \mathbf{M}_T^l = \mathbb{f}_T^{l}(\hat{\mathbf{M}}_T^{l-1})\\
\end{array} 
\right. ,
\label{encoder_t}
\end{equation}
where $\mathbb{f}_T^l$ denotes a GCN-based encoder to model the temporal dependency of $\hat{\mathbf{X}}_T$ and generate the temporal representation $\mathbf{M}_T^l \in \mathbb{R}^{(T_p+T_f) \times C}$. 
A one-layer GCN is then employed as the decoder $G_T$, yielding the predictive result as follows:
\begin{equation}
    \mathbf{Y}^{T,l} = \textrm{Reshape}\left(G_T(\mathbf{M}_T^l)\right) + \mathbf{X},
    \label{edcoder_t}
\end{equation}
where $\textrm{Reshape}(\cdot)$ is the operation to reformat the shape of data arrays and $\mathbf{Y}^{T,l}\in \mathbb{R}^{(T_p+T_f) \times J \times D}$.

\noindent\textbf{Spatial Domain Modeling.}
Given $\mathbf{X}$, we reshape it into the spatial domain as $\mathbf{X}_S = \{ X_{S,n} \}_{n=1}^{J \times D}\in \mathbb{R}^{(J \times D) \times (T_p+T_f)}$, where $X_{S,n}$ denotes a joint's coordinate trajectory along the time. We then apply discrete cosine transform and a joint embedding to obtain the joint representation $\hat{\mathbf{X}}_S = \{\hat {X}_{S,n} \}_{n=1}^{J\times D} \in \mathbb{R}^{(J\times D) \times C}$. 
To capture the spatial dependencies, we treat the human pose as a generic graph and use GCN to obtain the joint representation $\mathbf{M}_S$. $\mathbf{M}_S$ serves as input to the spatial domain decoder to generate the prediction $\mathbf{Y}^{S}$.
\begin{equation}
\left \{
\begin{array}{lll}
l=1 : \mathbf{M}_S^l = \psi_S^l(\hat{\mathbf{X}}_S) \\
l>1 : \mathbf{M}_S^l = \psi_S^l(\hat{\mathbf{M}}_S^{l-1})\\
\end{array} 
\right.
,
\label{encoder_s}
\end{equation}

\begin{equation}
    \mathbf{Y}^{S,l} = \textrm{Reshape}\left( \textrm{IDCT}\left(G_S(\mathbf{M}_S^l)\right) \right)  + \mathbf{X},
    \label{edcoder_s}
\end{equation}
where $\psi_S^l$ and $G_S$ denote the GCN-based encoder and decoder of the spatial domain, respectively. The inverse discrete cosine transform, $\textrm{IDCT}(\cdot)$, recovers the pose space. $\mathbf{Y}^{S,l} \in \mathbb{R}^{(T_p+T_f) \times J \times D} $ represents the prediction of the $l$-th layer.

\noindent\textbf{Spatio-Temporal Information Interaction.}
Knowledge learned from each domain complements each other, and cross-domain knowledge distillation facilitates information exchange, reducing bias in the learned representations. 
We utilize Mean Per Joint Position Error (MPJPE) as a constraint to encourage knowledge transfer between domains, enabling both branches to consider the distinctive information related to temporal smoothness and spatial dependencies. 
\begin{equation}
    \mathcal{L}_{ST} = \sum_{l=1}^{L} \frac{1}{(T_p+T_f)\cdot J} \sum_{t=1}^{T_p+T_f} \sum_{j=1}^{J} \parallel \mathbf{Y}^{T,l}_{t,j} - \mathbf{Y}^{S,l}_{t,j} \parallel^2.
    \label{Eq:stc}
\end{equation}
Through Eq.\eqref{Eq:stc}, the dual-branch model enables reciprocal knowledge transfer during training. The proposed loss function supports direct consistent constraints across domains at multiple scales, promoting continuous knowledge exchange. 
Given GCN's strength in modeling joint motion correlations and the discrete cosine transform's ability to capture joint space trajectories, we take the output of the spatial domain branch as the final result, denoted by $\widetilde{\mathbf{Y}} = \mathbf{Y}^{S,L}$.

\subsection{Multi-Subgraph Learning} 
Given $\hat{\mathbf{X}}_T$ $(l=1)$ or $\mathbf{M}_T^l$ $(l>1)$, where $T$ denotes the temporal domain, we employ multi-graph convolutions for feature learning. Let $\mathbb{f}_T^l=\{\mathbb{f}_{T}^{l,1},\mathbb{f}_{T}^{l,2},\cdots,\mathbb{f}_{T}^{l,K}\}$, where $K$ is the number of graph convolution kernels. The calculation for each kernel is defined as $H_{T}^{l,k} = \sigma (A_{T}^{l,k} \hat{\mathbf{X}}_T W_{T}^{l,k})$ or $H_{T}^{l,k} = \sigma (A_{T}^{l,k} \mathbf{M}_T^l W_{T}^{l,k})$, where $k=\{1,2,\cdots, K\}$, $A_{T}^{l,k} \in \mathbb{R}^{(T_p+T_f) \times (T_p+T_f)}$ and $W_{T}^{l,k}\in \mathbb{R}^{C \times C}$ denote the adjacency matrix and the trainable transformation matrix, respectively. The output of $\mathbb{f}_T^l$ can be expressed as:
\begin{equation}
    \mathbf{M}_T^l = \textrm{Ave}(H_{T}^{l,1},H_{T}^{l,2},\cdots,H_{T}^{l,K}),
\end{equation}
where $\textrm{Ave}(\cdot)$ denotes the averaging operation. The different kernels share the same objective, effectively partitioning the problem space into homogeneous regions. To prevent excessive divergence among kernels, we propose a homogeneous information learning enhancement strategy. This involves constraining the learning of adjacency matrices for each kernel to ensure similar information aggregation among different convolution kernels. Specifically, we compare adjacency matrices directly during training to enforce this constraint.


\begin{equation}
    \mathcal{L}_{con}^{T} = \sum_{l=1}^L \sum_{k=1}^{K} \sum_{u=k+1}^{K}\parallel A_{T}^{l,k} - A_{T}^{l,u}\parallel^2_2.
\end{equation}
Similarly, we have the constraint in the spatial domain:
\begin{equation}
    \mathcal{L}_{con}^{S} = \sum_{l=1}^L \sum_{k=1}^{K} \sum_{u=k+1}^{K}\parallel A_{S}^{l,k} - A_{S}^{l,u}\parallel^2_2,
\end{equation}
where $A_{S}^{l,k} \in \mathbb{R}^{(J\cdot D ) \times (J\cdot D)}$ represents the adjacency matrix in the spatial domain.

\subsection{Loss Function}
For training the proposed model, we use the average $\ell_2$ distance between predicted and ground-truth joint positions as the loss $\mathcal{L}_1$. Formally, for one sample, $\mathcal{L}_1$ is defined as:
\begin{equation}
    \mathcal{L}_1 = \frac{1}{(T_p+T_f)\cdot J} \sum_{t=1}^{T_p+T_f} \sum_{j=1}^{J} \parallel \widetilde{\mathbf{Y}}_{t,j} - \mathbf{Y}_{t,j} \parallel^2 .
\end{equation}
The final loss function $\mathcal{L}$ combines $\mathcal{L}_1$ with $\mathcal{L}_{ST}$, $\mathcal{L}_{con}^{S}$, and $\mathcal{L}_{con}^{T}$, weighted by a hyperparameter $\lambda$.
\begin{equation}
    \mathcal{L} = \mathcal{L}_1 +  \lambda \cdot \left( \mathcal{L}_{ST} + \mathcal{L}_{con}^{S} + \mathcal{L}_{con}^{T}\right).
\end{equation}

\section{Experiments}
\label{sec:experiment}

\subsection{Datasets and baselines}
We evaluate our method on the Human3.6m (H3.6M)~\cite{ionescu2013human3} and CMU Motion Capture (CMU Mocap)~\cite{dang2021msr,li2022skeleton} datasets,  comparing it with Traj-GCN~\cite{mao2019learning}, DMGNN~\cite{li2020dynamic}, 
STSGCN~\cite{sofianos2021space}, MSR-GCN~\cite{dang2021msr}, SPGSN~\cite{li2022skeleton}, PGBIG~\cite{ma2022progressively}, 
and STBMP without incremental information~\cite{wang2023spatio}. Mean Per Joint Position Error (MPJPE) is reported to evaluate the performance, the lower indicates better predictive performance.

\begin{table}[!t]\scriptsize
\centering
\setlength{\tabcolsep}{7.5pt} 
\caption{Comparisons of average MPJPEs across all actions in H3.6M. 
Red/blue font indicates the best/second best result.
}
\label{tab:h3.6m}
\begin{tabular}{lcccccc}
\toprule
 \multirow{1}{*}{Mothod}  & 80ms & 160ms & 320ms & 400ms & 560ms & 1000ms\\ \hline
Traj-GCN & 12.68 & 26.06 & 52.27 & 63.51 & 81.07 & 113.01\\
DMGNN    & 16.95 & 33.62 & 65.90 & 79.65 & 93.57 & 127.62\\
MSR-GCN  & 12.11 & 25.56 & 51.64 & 62.93 & 81.13 & 114.18\\
STSGCN   & 15.34 & 25.52 & 50.64 & 60.61 & 80.66 & 113.33\\
SPGSN  & 10.44 & \textcolor{blue}{22.33} & \textcolor{blue}{47.07} & \textcolor{blue}{58.26} &\textcolor{blue}{77.40} & \textcolor{blue}{109.64}\\
PGBIG &10.33 &22.74 &47.45 &58.47 &\textcolor{red}{76.91} 	&110.31\\
STBMP &\textcolor{blue}{9.98} &22.45 &47.77 &59.06 &80.45 &112.56\\
Ours &\textcolor{red}{9.61} &\textcolor{red}{21.63} &\textcolor{red}{46.40} &\textcolor{red}{57.55} &77.81 &\textcolor{red}{109.51}\\ \bottomrule
\end{tabular}
\vspace{-0.1in}
\end{table}

\subsection{Experiment Results}
\noindent\textbf{Motion prediction.}
Tab.~\ref{tab:h3.6m} and Tab.~\ref{table:CMU} show the prediction performance of each method at different future time steps.
Our method outperforms the baselines in most cases, $e.g.$, an average reduction of 3.71\% at 80ms and 3.13\% at 160ms on H3.6M. These results prove the effectiveness of our method in achieving improvements in motion prediction.
Besides, to qualitatively evaluate the prediction, we also visualize some videos to compare the predicted poses by baselines and our method in the external link~\footnote{https://github.com/JasonWang959/STMS}. The results show that STMS-GCN generates more precise and reasonable future poses.

\begin{table}[!t]\scriptsize
\centering
\setlength{\tabcolsep}{7pt} 
\caption{Comparison of average MPJPEs across all actions on the CMU-Mocap dataset. 
}
\vspace{-0.05in}
\label{table:CMU}
\begin{tabular}{l|ccccccc}\toprule
 Method & 80ms & 160ms & 320ms & 400ms  & 1000ms &Average \\ \hline 
 Traj-GCN & 9.94 &18.02 &33.55 &40.95 &81.85 &36.86\\
 DMGNN    & 14.07 &24.44 &45.90 &55.45 &104.33 &39.35\\
 MSR-GCN  & 8.72 &15.83 &30.57 &38.10 &79.01 &34.45\\
 STSGCN   & 10.80 &18.19 &31.18 &41.05&81.76 &36.60\\
 SPGSN    & 8.30 &\textcolor{blue}{14.80}& \textcolor{blue}{28.64} &\textcolor{blue}{36.96} &77.82 &\textcolor{blue}{33.30}\\
 PGBIG &  \textcolor{blue}{8.20} &15.41 &30.13 &37.27 &\textcolor{blue}{76.69} &33.54\\
 STBMP & 8.22 &15.33 &30.53 &37.95 &77.27 &33.86 \\
Ours & \textcolor{red}{7.70} &\textcolor{red}{14.04} &\textcolor{red}{28.57} &\textcolor{red}{36.03}  &\textcolor{red}{75.81} & \textcolor{red}{32.43}
 \\ \bottomrule
\end{tabular}
\vspace{-0.1in}
\end{table}

\noindent\textbf{Ablation of STMSB Block Design.} 
We explore the effect of the proposed modules in our method on H3.6M, with results summarized in Tab.~\ref{table:ablation_study}. It is observed that: 1) the proposed modules all contribute to an accurate prediction; and 2) the the full model, incorporating all components, yields the best performance. This underscores the collaborative benefits of integrated components.
Furthermore, in the full model, the average predictive performance of $\mathbf{Y}^{T,L}$ and $\frac{1}{2}(\mathbf{Y}^{T,L}+\mathbf{Y}^{S,L})$ are 34.52 and 33.98, respectively, supporting the choice of $\mathbf{Y}^{S,L} (33.80)$ as the final prediction.

\begin{table}[!t]\scriptsize
\setlength{\tabcolsep}{3pt} 
\centering
\caption{Ablation results for our methods. $\mathbf{G}_S$ and  $\mathbf{G}_T$ denote the spatial and temporal domain modeling, respectively.}
\vspace{-0.05in}
\label{table:ablation_study}
\begin{tabular}{ccccc|ccccc}
\toprule
$\mathbf{G}_S$ &$\mathbf{G}_T$ &$\mathcal{L}_{con}$ &$\mathcal{L}_{ST}$ &$\mathcal{L}_1$ & 80ms &160ms &320ms &400ms & Average\\ \hline 
$\checkmark$& & & &$\checkmark$ & \textcolor{red}{9.58}  &22.05 &48.84  &60.79 &35.32\\
$\checkmark$& &$\checkmark$ & &$\checkmark$ & 9.88 &22.17 &48.07 &59.61 &34.93\\
&$\checkmark$ & & &$\checkmark$ & 10.68 &23.60 &49.73 &61.42 &36.36\\ 
&$\checkmark$ &$\checkmark$ & &$\checkmark$ & 10.56 &23.29 &49.19  &60.89 &35.98\\ 
$\checkmark$ &$\checkmark$ & & $\checkmark$ &$\checkmark$ & 9.83 &\textcolor{blue}{21.95}  &\textcolor{blue}{47.50} &\textcolor{blue}{59.08} &\textcolor{blue}{34.59}\\ 
$\checkmark$ & $\checkmark$ & $\checkmark$ & $\checkmark$  &$\checkmark$ & \textcolor{blue}{9.61} &\textcolor{red}{21.63} &\textcolor{red}{46.40} &\textcolor{red}{57.55} &\textcolor{red}{33.80}\\ 
\bottomrule  
\vspace{-0.15in}
\end{tabular}
\end{table}

\begin{figure}[!t]
\vspace{-0.1in}
\centering
\begin{minipage}{0.43\linewidth}
\centering
\setlength{\abovecaptionskip}{0.28cm}
\includegraphics[width=0.99\linewidth]{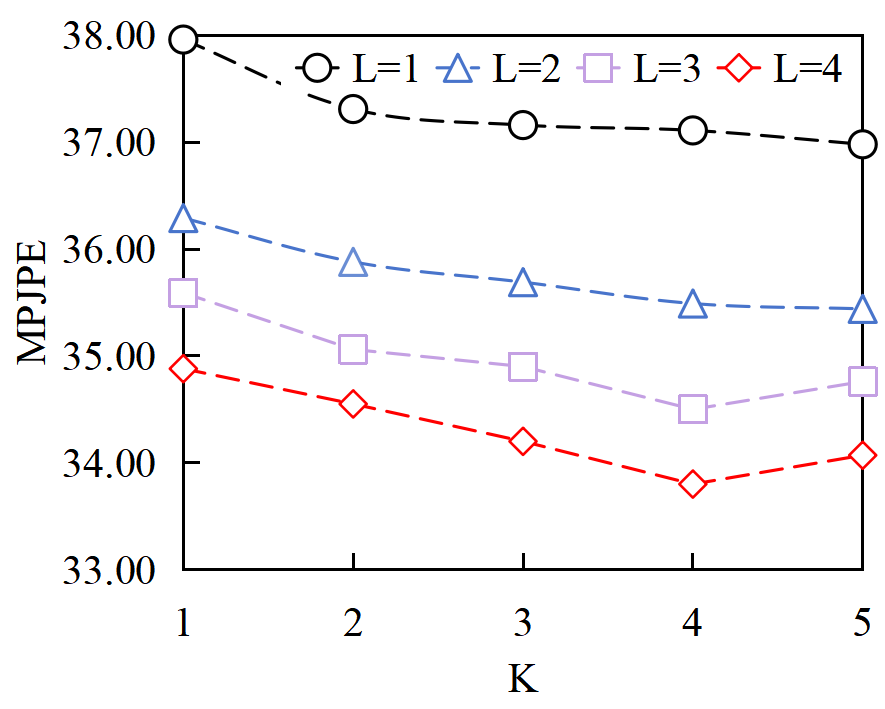}
\vspace{-0.2in}
\caption{Comparison of the predictive performance.}
    \label{fig:K_L}
\vspace{-0.1in}
\end{minipage}
\hfill
\begin{minipage}{0.48\linewidth}
\setlength{\tabcolsep}{1pt}
\caption{Impact of the fusion coefficient $\lambda$ on the motion predictive performance.}
\label{table:lambda}
\resizebox{\linewidth}{!}{
\begin{tabular}{l|ccccc}
\toprule
$\lambda $ &80ms &160ms &320ms &400ms &Avg.\\ \hline
 0     & 9.69 & 22.06 & 48.44 & 60.24 & 35.11\\
 0.001 & 9.80 &21.98 &47.63 &59.23 &34.66\\ 
 0.01  & \textcolor{blue}{9.61} &\textcolor{blue}{21.75} &\textcolor{blue}{47.26} &\textcolor{blue}{58.65} &\textcolor{blue}{34.32}\\ 
 0.1   & \textcolor{red}{9.61} &\textcolor{red}{21.63} &\textcolor{red}{46.40} &\textcolor{red}{57.55} &\textcolor{red}{33.80}\\  
 1.0   & 10.83 &24.71 &52.58 &64.46 &38.15\\ 
 \bottomrule
\end{tabular}}
\end{minipage}
\vspace{-0.15in}
\end{figure}

\noindent\textbf{Effect of STMSBs ($L$) and graph convolutions ($K$).} 
Fig.\ref{fig:K_L} shows the average MPJPEs of short-term prediction (80, 160, 320, and 480 ms) for different architectures on H3.6M. The results show that increasing $K$ and $L$ generally improves performance. Experimentally, we chose $L=4$ and $K=4$ throughout this paper.
Besides, we can find that using only one graph convolution ($K=1$) generally performs worse than the settings with larger $K$, where the latter enables the model to discover more potential motion patterns. These results further highlight the advantage of our design.

\noindent\textbf{Influence of hyper-paramter $\lambda$.} 
We systematically varied the hyper-paramter $\lambda$ across $\{0, 10^{-3}, 10^{-2}, 10^{-1}, 1\}$ to assess its impact on H3.6M. As shown in Tab.~\ref{table:lambda}, our approach exhibits an average performance improvement of 1.28\% on $\lambda=0.001$, 2.25\% on $\lambda=0.01$, 3.73\% on $\lambda=0.1$, highlighting the benefits of cross-domain and homogeneous information constraint. However, performance decreases by 8.66\% at $\lambda=1.0$, suggesting that excessive constraints may limit the uniqueness of spatio-temporal branch information and reduce the effectiveness of diverse graph convolutions.

\begin{figure}[!t]
\centering
\begin{minipage}{0.49\linewidth}\centering
\setlength{\tabcolsep}{1pt} 
\captionof{table}{Dfferent consistency constraints in the multi-subgraph learning are applied to weight parameters ``W'' or adjacency matrices ``A''. 
}
\vspace{-0.05in}
\resizebox{\linewidth}{!}{
\begin{tabular}{l|cccccc}
\toprule 
Constraint &80ms &160ms &320ms &400ms &Avg. \\ \hline
$\times$ & 9.83 &21.95  &47.50 &59.08 &34.59\\
W &  \textcolor{blue}{9.66} &21.77 &46.66 &57.81 &33.98\\
A (Ours) & \textcolor{red}{9.61} &\textcolor{red}{21.63} &\textcolor{red}{46.40} &\textcolor{red}{57.55} &\textcolor{red}{33.80} \\ 
W and A & 9.68 &\textcolor{blue}{21.72} &\textcolor{blue}{46.53} &\textcolor{blue}{57.72} &\textcolor{blue}{33.91} \\ 
 \bottomrule
\end{tabular}}
\label{tab:wa}
\end{minipage}
\hfill
\begin{minipage}{0.47\linewidth}\centering
\setlength{\tabcolsep}{1pt} 
\captionof{table}{Impact of consistency. We vary the coefficient $\beta$ under $\mathcal{L} = \mathcal{L}_1 + 0.1\cdot \mathcal{L}_{ST} + \beta \cdot \left( \mathcal{L}_{con}^{S} + \mathcal{L}_{con}^{T}\right)$.
}
\vspace{-0.05in}
\resizebox{\linewidth}{!}{
\begin{tabular}{l|cccccc}
\toprule
$\beta $ &80ms &160ms &320ms &400ms &Avg. \\ \hline
 -1.0    & 9.94 &22.34 &47.74 &59.13 &34.79\\
 -0.1  & 9.86 &22.19 &47.56 &58.90 &34.63\\ 
 0     & 9.83 &21.95  &47.50 &59.08 &34.59\\
 0.1   & \textcolor{red}{9.61} &\textcolor{red}{21.63} &\textcolor{red}{46.40} &\textcolor{red}{57.55} &\textcolor{red}{33.80}\\  
 1.0   & \textcolor{blue}{9.64} &\textcolor{blue}{21.68} &\textcolor{blue}{46.50} &\textcolor{blue}{57.70} &\textcolor{blue}{33.88}\\ 
 \bottomrule
\end{tabular}}
\label{tab:cvsd}
\end{minipage}
\vspace{-0.2in}
\end{figure}

\noindent\textbf{Effect of the homogeneous information constrain.}
In multi-subgraph learning, constraining the adjacency matrix or the trainable weights ensures similar information aggregation or information mapping across convolution kernels. Tab.~\ref{tab:wa} shows that considering homogeneous information constraints consistently improves model performance, with enforcing graph construction similarity alone yielding the best. 
The adjacency matrix affects node connectivity and information propagation. By constraining adjacency matrix similarity, we ensure a more consistent and stable information propagation. However, overly strict constraints may limit the flexibility of weight parameters and the model's capacity to capture diverse information.
To further investigate, we compare consistency ($\beta>0$) and diversity constraints ($\beta<0$) of adjacency matrices. 
Tab.~\ref{tab:cvsd} indicates that enforcing similarity in graph construction has better performance and vice versa, highlighting the importance of consistency in subgraph learning.

%

\section{Conclusion}
\label{sec:conclusion}
In this paper, we propose a novel spatio-temporal multi-subgraph graph convolutional network for human motion prediction. Different from previous approaches, our method comprehensively considers the independence and complementarity of spatio-temporal information in human motion. By separately modeling temporal and spatial features, and employing consistency constraints on intermediate spatio-temporal features, we enable effective capture and utilization of intricate spatio-temporal relationships. Additionally, multi-subgraph learning extracts more expressive motion patterns, and a homogeneous information constraint enhances subgraph learning. Extensive experiments on the human motion prediction benchmarks validate the effectiveness of our method. 

\section*{Acknowledgment}
\addcontentsline{toc}{section}{Acknowledgment}
This work was supported in part by the National Natural Science Foundation of China No. 62376277 and Public Computing Cloud, Renmin University of China.

\bibliographystyle{IEEEtran}
\bibliography{IEEEabrv}

\end{document}